\PassOptionsToPackage{table}{xcolor}
\documentclass[letterpaper,twocolumn,10pt]{article}
\usepackage{lineno}
\usepackage{hegroup}
\usepackage{enumitem}
\usepackage{multirow}
\usepackage{tikz}
\usepackage[breakable]{tcolorbox} 
\tcbuselibrary{most,listings,breakable}
\usepackage{amsfonts}
\usepackage{xspace} 
\usepackage{amsmath}
\usepackage{listings}
\usepackage{mathrsfs}
\usepackage{amssymb}
\usepackage{amsmath}
\usepackage{amsthm}
\usepackage[ruled,vlined]{algorithm2e} 
\usepackage[normalem]{ulem}
\useunder{\uline}{\ul}{}
\usepackage{makecell}
\usepackage{caption}
\usepackage{subcaption}
\usepackage{float}
\usepackage{booktabs}
\usepackage{balance}
\usepackage{xcolor}
\usepackage{tcolorbox}
\usepackage{xurl}
\usepackage{hyperref}
\hypersetup{
  colorlinks,
  linkcolor={blue!70!green},
  citecolor={green!70!blue},
  urlcolor={orange!70!red}
}
\usepackage{cleveref}

\usepackage[hang,flushmargin]{footmisc}

\newtcolorbox[auto counter,number within=section]{pabox}[2][]{%
colback=green!10,colframe=green!50,fonttitle=\bfseries,coltitle=black, 
title=Summary.~\thetcbcounter}
\newcommand{\Method}{$\mathsf{ThoughtMani}$\xspace}

\definecolor{maincolor}{HTML}{1f77b4} 
\definecolor{highlight}{HTML}{4169E1} 

\newtcolorbox{cotbox}[1][]{
    colback=maincolor!10,
    colframe=maincolor,
    width=\columnwidth,
    fonttitle=\bfseries,
    coltitle=white,
    arc=1mm,
    auto outer arc,
    left=4pt,
    right=4pt,
    breakable,
    title=#1,
}

\newtcbox{\highlightbox}[1][]{
    colback=highlight!20,
    colframe=highlight!80,
    boxrule=0.5pt,
    arc=1mm,
    outer arc=1mm,
    left=2pt,
    right=2pt,
    top=1pt,
    bottom=1pt,
    #1
}
\newcommand{\mypara}[1]{\smallskip\noindent{\bf {#1}.}}

\begin{document}

\date{}
\title{Thought Manipulation: External Thought Can Be Efficient for Large Reasoning Models}

\author{
Yule Liu\textsuperscript{1}  \ \ \ 
Jingyi Zheng\textsuperscript{1}  \ \ \ 
Zhen Sun\textsuperscript{1}  \ \ \ 
Zifan Peng\textsuperscript{1}  \ \ \ 
Wenhan Dong \textsuperscript{1}  \ \ \ \\
Zeyang Sha \textsuperscript{2} \ \ \
Shiwen Cui \textsuperscript{2} \ \ \
Weiqiang Wang \textsuperscript{2} \ \ \
Xinlei He\textsuperscript{1}\thanks{Corresponding author (\href{mailto:xinleihe@hkust-gz.edu.cn}{xinleihe@hkust-gz.edu.cn}).} \ \ \ 
\\
\\
\textsuperscript{1}\textit{Hong Kong University of Science and Technology (Guangzhou)} \ \ \ 
\textsuperscript{2}\textit{Ant Group} \ \ \ 
}

\maketitle

\begin{abstract}

Recent advancements in large reasoning models (LRMs) have demonstrated the effectiveness of scaling test-time computation to enhance reasoning capabilities on various tasks.
However, LRMs often suffer from an ``overthinking'' problem, where the model generates excessively redundant reasoning steps with limited performance gains.
In this work, we empirically reveal an important characteristic of LRM behaviors that placing external CoTs generated by smaller models between the thinking token (\texttt{<think>} and \texttt{</think>}) can effectively manipulate the model to generate fewer thoughts.
Building on this finding, we propose a simple yet efficient pipeline, \Method, to enable LRMs to bypass unnecessary intermediate steps, thereby significantly reducing computational costs. 
We conduct extensive experiments to evaluate the utility and efficiency of \Method.
For instance, when applied to QwQ-32B on the LiveBench/Code dataset, \Method keeps the original performance while reducing output token counts by approximately 30\%, with minimal overhead introduced by the CoT generator.
Furthermore, we identify two suboptimal modes, blindly following flawed external thoughts and unnecessary rethinking, and show that simple mitigations, such as difficulty-aware fallbacks, can further improve performance. 
Overall, \Method offers a practical, general, and efficient way to optimize LRM inference, making powerful reasoning models more accessible and scalable for real-world applications.

\end{abstract}

\section{Introduction}

Recent advancements in large reasoning models (LRMs) have demonstrated the great potential of incorporating long-thinking processes in enhancing reasoning capabilities for complex tasks~\cite{plaat2024reasoning,xu2025towards}. 
By leveraging reinforcement learning (RL), LRMs are trained to generate step-by-step chain-of-thought (CoT) reasoning, breaking down problems into smaller components and performing multiple checks before arriving at a final response~\cite{openai2025o1,team2025kimi,shao2024deepseekmath}.
Models like DeepSeek-R1~\cite{DBLP:journals/corr/abs-2501-12948} and QwQ~\cite{qwq32b} exemplify the effectiveness of this method, showcasing substantial improvements in reasoning accuracy.

Despite the improved performance, scaling up CoT often requires exponentially larger computational resources~\cite{sui2025stop}. 
Models like QwQ consume 5 to 10 times more tokens to reach conclusions compared to standard approaches~\cite{chen2024not}. 
Previous studies introduce ``overthinking'' to describe the phenomenon that unnecessary steps may lead to inefficiencies, particularly in simple questions~\cite{chen2024not}.
This inefficiency not only undermines the utility of LRMs in time-sensitive scenarios but also leads to additional computational resource consumption, further degrading overall system performance.

In this work, we first investigate a natural yet underexplored question arising from the inefficiencies of the inherent thinking process in LRMs: \textbf{Can externally provided thoughts substitute and bypass the model’s internal reasoning?} 
To answer this, we design a thought-replacing strategy that simulates the standard inference process, but with the internal CoT replaced by an external one generated by a smaller model. 
This external CoT is inserted between the \texttt{<think>} and \texttt{</think>} tokens in the input prompt, allowing us to observe whether the LRM accepts the provided reasoning and skips redundant thinking. 

Our analysis shows that LRMs can indeed skip their internal reasoning process and proceed directly to answer generation when presented with external thoughts. 
Notably, RL-based LRMs exhibit a nuanced behavior: while they often accept the external CoT, they sometimes resume thinking, indicated by the reappearance of the \texttt{</think>} token in their output—a behavior we term "rethinking." 
In contrast, distillation-based LRMs almost always terminate reasoning upon encountering \texttt{</think>}, leading to near-complete skipping. 
Further hidden-state-level analysis, based on the logit dynamics of the \texttt{</think>} token, confirms that this skipping behavior is not merely a superficial pattern match but reflects deeper internal state transitions. 
The rank of \texttt{</think>} in the output distribution drops sharply when the model decides to exit reasoning, suggesting a genuine shift in the model’s internal decision-making process. 

Crucially, we find that skipping internal reasoning leads to substantial efficiency gains, often reducing output token counts by over 50\%, while incurring only moderate performance drops (under 10\%). 
Building on these insights, we propose \Method, an efficient, training-free inference pipeline that leverages external CoTs to reduce computation.

We comprehensively evaluate ten different compression techniques across three LRMs on four diverse datasets. 
For example, when using Qwen-2.5-7B-Instruct as the CoT generator for QwQ-32B on the GSM-8k dataset, \Method reduces output token count by approximately 40\% (from 1,791 to 1,075 tokens), with an average overhead of only 52 additional tokens from the external thoughts. 
We also identify and quantify the frequency of two common suboptimal behaviors that affect efficiency.
By applying a simple difficulty-aware mitigation, excluding external thoughts for the most challenging problems, we improve accuracy by up to 8\% while still achieving substantial token savings compared to the full reasoning baseline.
Further ablation studies confirm the robustness and practical effectiveness of \Method across different settings and model types. 

Our contribution can be summarized as follows:
\begin{itemize}[leftmargin=*,itemsep=0.1em, topsep=0.2em]
    \item We reveal a unique pattern of LRMs when external thoughts are given, which sheds light on LRMs' characteristics. 
    Specifically, we uncover how RL-based and distillation-based LRMs differ in their handling of provided CoT, offering insights into their reasoning mechanisms and decision-making processes. 
    \item We introduce \Method, a training-free inference pipeline that reduces redundant reasoning by injecting thoughts generated from smaller models, achieving substantial computational savings without compromising performance or needing additional training.
    
    \item Extensive experiments demonstrate the superiority of \Method in both utility and efficiency, with consistent improvements across diverse datasets and tasks, underscoring its practical value in real-world scenarios.
\end{itemize}

\begin{figure*}[t]
        \centering
    \includegraphics[width=0.33\textwidth]{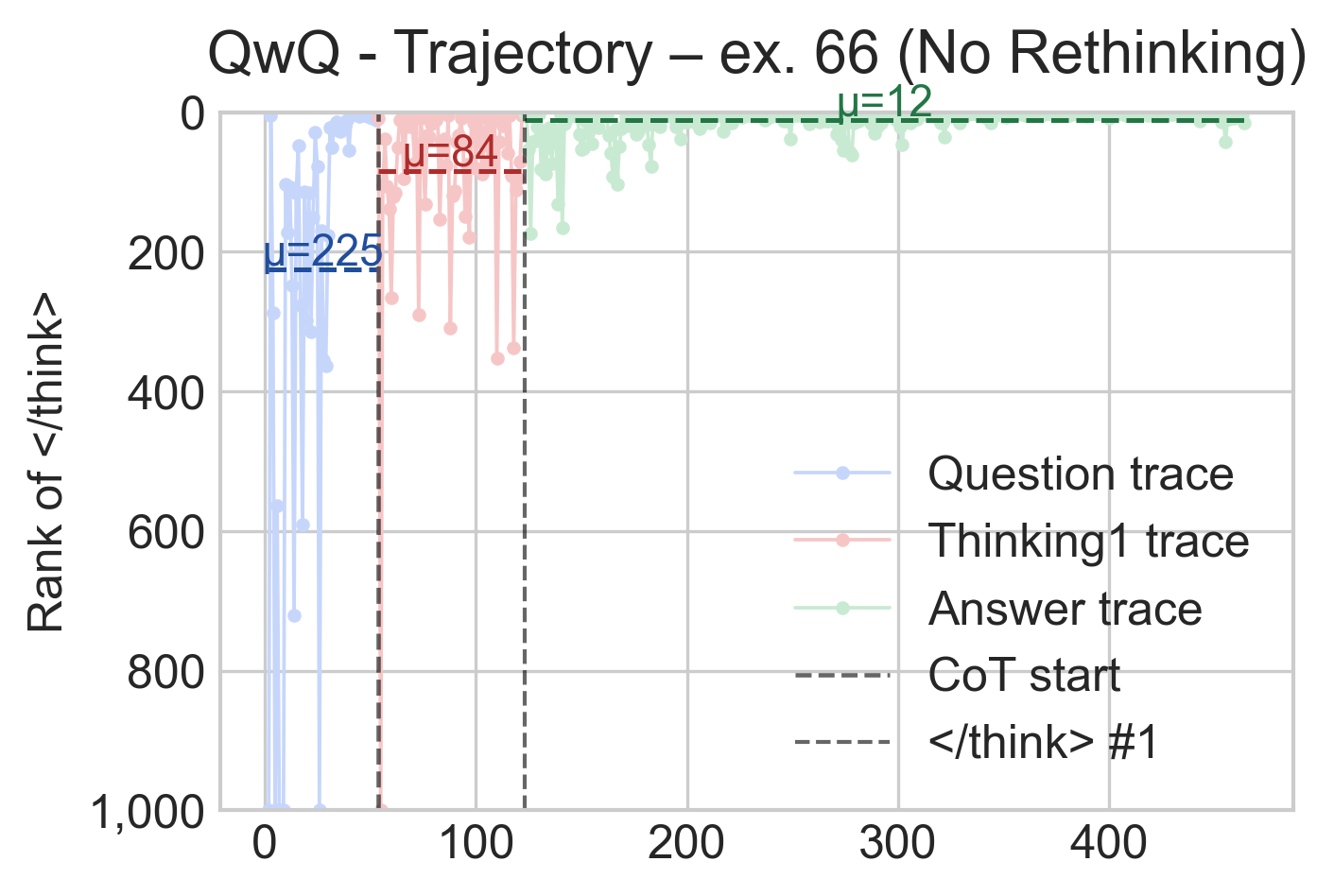}
    \includegraphics[width=0.33\textwidth]{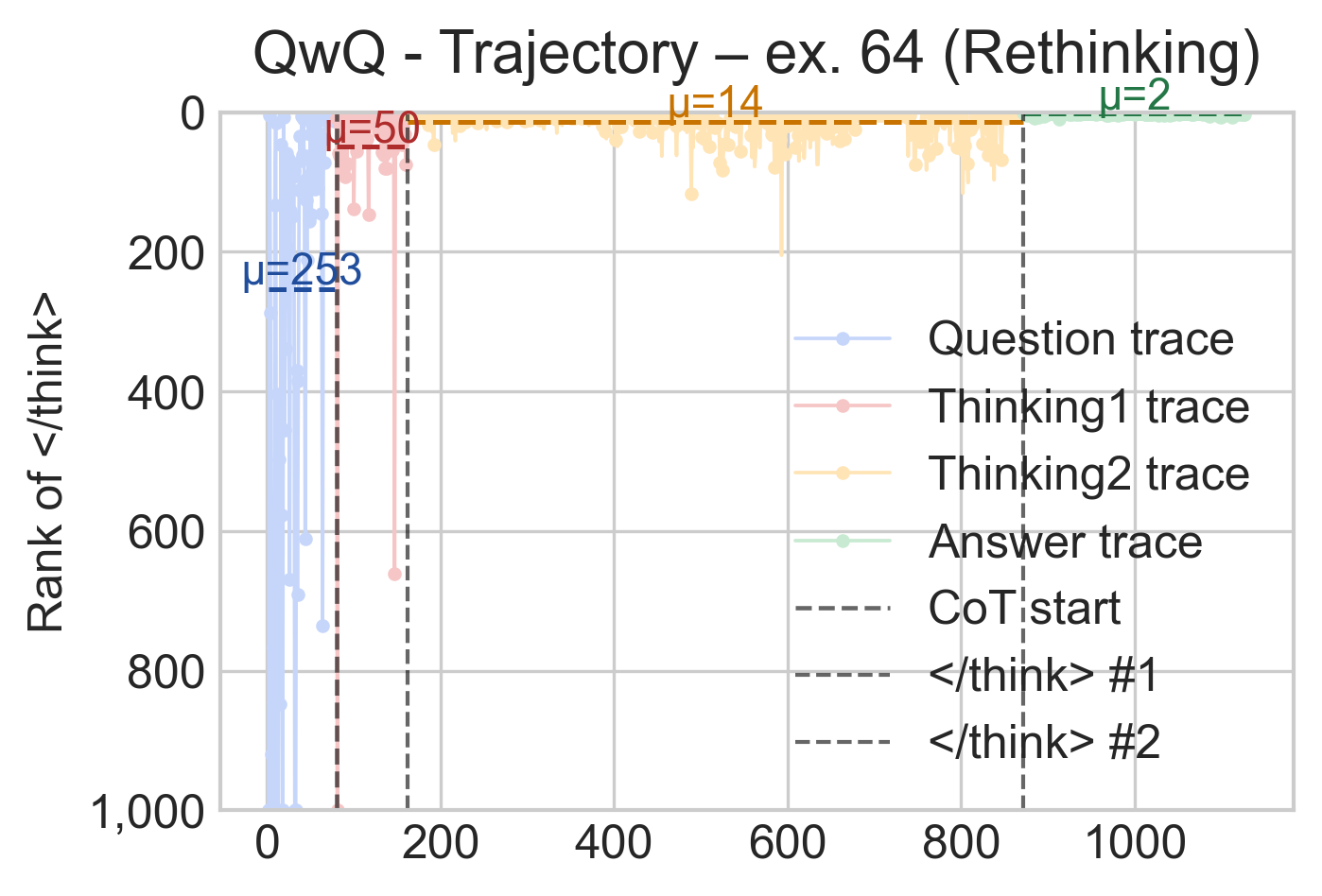}
    \includegraphics[width=0.33\textwidth]{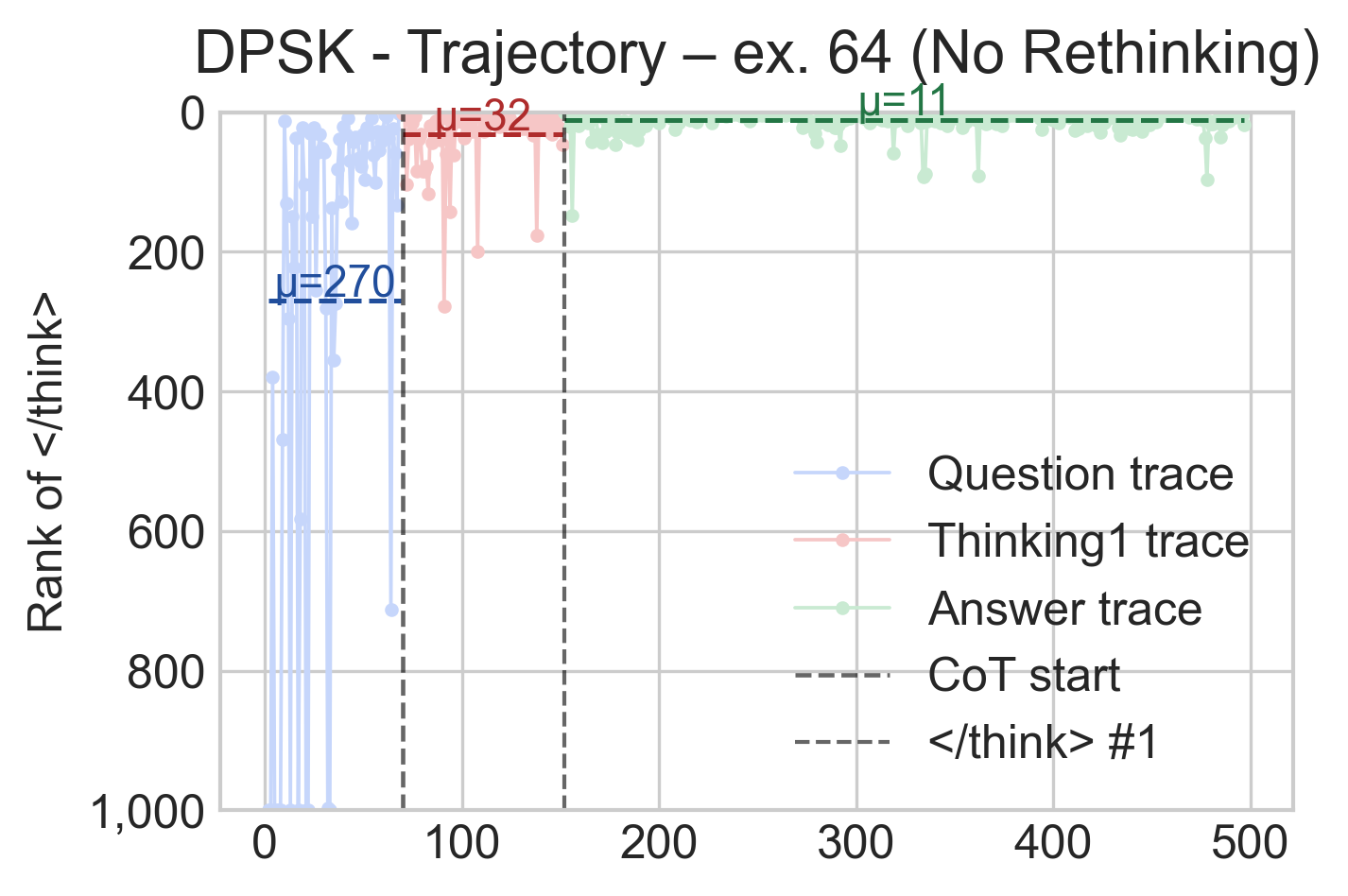}
    \caption{\textbf{Rank Trajectory of \texttt{</think>}}: For a prompt and answer pair, we identify the question, thinking, and answer traces by corresponding special tokens, i.e., \texttt{<|im\_end|>} and \texttt{</think>}.
    The CoT generator is Qwen-Plus, and the inference models are QwQ and Deepseek-Distillation-Qwen-2.5-32B-Instruct (DPSK), respectively.
    Specifically, each trace is assigned to a different color and marked with the mean rank value $\mu$ and the rank is cut off by 1000 for clear demonstration.
    The content of these trajectories is shown in the supplementary materials.
    }
        \label{fig:trajectory}

\end{figure*}

\section{Effectiveness of External Thoughts}
\label{sec:obers}

In this section, we try to answer the question:
\textbf{Can externally provided thoughts substitute and bypass the generation of the original ones?}
First, we investigate the effect of placing external thoughts in positions typically reserved for model-generated thoughts, revealing distinct behaviors between RL-based and distillation-based LRMs in terms of skipping the original reasoning process.
Second, we perform a fine-grained analysis of the remaining tokens to understand how external thoughts influence the reasoning process and token allocation.
Finally, based on these insights, we introduce \Method, a training-free framework designed to enable efficient inference by leveraging external thoughts. 

\begin{table}[t!]
\centering
\caption{\textbf{Statistics of \texttt{</think>} Token Rank}:  
Mean position (± std) of the \texttt{</think>} token in different response stages.  
Only samples exhibiting rethinking (i.e., where \texttt{</think>} reappears) are included.  
Columns correspond to Question, External Chain-of-Thought (CoT), Rethinking (Re.), and Final Answer stages.}
\label{tab:stoptime}
\resizebox{\linewidth}{!}{
\begin{tabular}{c|c|c|c|cccc}
\toprule
\multicolumn{2}{c|}{\textbf{Ex.}} & \textbf{Skip?} & \textbf{Num.} & \textbf{Question} & \textbf{CoT} & \textbf{Re.} & \textbf{Answer} \\
\midrule
\multirow{10}{*}{\rotatebox{90}{\textbf{QwQ}}}
& \multirow{2}{*}{3B} &Yes  & 48  & 3278.0\textsubscript{19250.0} & 132.1\textsubscript{168.9} & -                & 12.7\textsubscript{23.6} \\
& &No   & 38  & 2715.0\textsubscript{17640.0} & 123.7\textsubscript{243.5} & 11.9\textsubscript{78.8} & 4.7\textsubscript{30.3} \\
\cmidrule{2-8}
& \multirow{2}{*}{7B}&Yes  & 93  & 2950.0\textsubscript{18402.0} & 66.8\textsubscript{142.4}  & -                & 15.3\textsubscript{30.4} \\
&  &No   & 130 & 2690.0\textsubscript{17554.0} & 76.5\textsubscript{175.0}  & 12.1\textsubscript{28.1} & 3.6\textsubscript{22.1} \\
\cmidrule{2-8}
& \multirow{2}{*}{Plus} &Yes& 281 & 2968.3\textsubscript{18456.8} & 69.6\textsubscript{125.2}  & -                & 18.5\textsubscript{35.0} \\
&  &No & 219 & 2592.7\textsubscript{17240.0} & 69.6\textsubscript{120.5}  & 13.4\textsubscript{57.6} & 3.1\textsubscript{5.5}  \\
\cmidrule{2-8}
& \multirow{2}{*}{Max} &Yes & 219 & 2899.5\textsubscript{18218.0} & 46.0\textsubscript{84.1}   & -                & 18.4\textsubscript{69.9} \\
&  &No  & 281 & 2686.0\textsubscript{17561.0} & 47.6\textsubscript{78.1}   & 13.0\textsubscript{36.0} & 6.1\textsubscript{78.9} \\
\midrule
\multirow{5}{*}{\rotatebox{90}{\textbf{Deepseek}}}
& 3B &Yes   & 86  & 5259.0\textsubscript{20324.0} & 97.8\textsubscript{162.1}  & - & 15.6\textsubscript{29.0} \\
\cmidrule{2-8}
& 7B &Yes   & 223 & 5261.3\textsubscript{20330.9} & 58.8\textsubscript{133.4}  & - & 13.2\textsubscript{25.7} \\
\cmidrule{2-8}
& Plus &Yes & 500 & 5263.5\textsubscript{20339.5} & 47.3\textsubscript{94.6}   & - & 12.4\textsubscript{23.7} \\
\cmidrule{2-8}
& Max &Yes  & 500 & 5262.0\textsubscript{20329.0} & 32.2\textsubscript{54.4}   & - & 11.3\textsubscript{21.2} \\
\bottomrule
\end{tabular}}
\end{table}

\subsection{Replacing the Original Thought}
We simulate the thinking process by placing external thoughts in positions typically reserved for model-generated thoughts.
To generate the external thoughts, we select Qwen series models~\cite{yang2024qwen2}, including Qwen-Max, Qwen-Plus, Qwen-2.5-7B-Instruct, and Qwen-2.5-3B-Instruct, to generate high-level ideas for solving the problems.
This selection enables a fair comparison of the generated thoughts, where the larger model typically exhibits stronger capability.
Specifically, the generation process is prompted to focus solely on high-level reasoning steps without delving into detailed calculations or producing final answers. 
The prompt is shown as follows:
\begin{cotbox}[Thought Generation]
``If you are a teacher, you are listing the important key points for solving the problem, and no calculation details should be included.
You are not allowed to produce any final answer. Add \texttt{<STOP>} when the key points are finished.
You may provide **only very high-level ideas** for solving the problem; no calculation details should be included. 
If you feel that you cannot solve it, output  \texttt{<STOP>} and return.''
\end{cotbox}

The generated thoughts are then enclosed within \texttt{<think>} and \texttt{</think>} tokens and inserted into a standardized chat template, as shown below: 
\begin{cotbox}[Inference Template]
    \texttt{<|im\_start|>} 
    \textbf{\color{maincolor} User:} \textbf{[Question]}
    \texttt{<|im\_end|>}

    \texttt{<|im\_start|>} 
    \textbf{\color{maincolor} Assistant:} 
    \texttt{<|im\_end|>}

    \texttt{<think>} 
    \textbf{\color{highlight} [Generated Thought]}
    \texttt{</think>}

\end{cotbox}

This placement simulates the normal inference process, except that the thought is now provided by an external source rather than being internally generated by the LRM. 
The template is then used to invoke LRMs such as QwQ and Deepseek-Distillation-Qwen-2.5-32b-instruct.
By analyzing the model's response, we evaluate whether the inherent thinking process is bypassed and how effectively the external thoughts are utilized.

\subsection{Skipped or Not}
In this part, we first empirically examine the skip of inherent thinking by the token-level feature in the prompts, and reveal distinct behaviors between RL-based and distillation-based LRMs.
Second, we perform a hidden-state-level analysis to uncover additional distinguishing features that further support the identification of thought skipping. 
The sample outputs are provided in the supplementary materials.

\mypara{Token-Level Analysis}
Since the external thoughts are included in the input prompt, we determine that the model skips its internal reasoning if the token \texttt{</think>} does not appear in the generated response. 
If \texttt{</think>} does appear, we consider the thinking process not skipped—meaning the LRM generates additional internal reasoning despite the provided external thought, which we refer to as "rethinking"\footnote{We use "not skipped" and "rethinking" interchangeably to describe this behavior.}. 
This empirical analysis is conducted on the MATH-500 dataset, a benchmark for evaluating reasoning capability.

The statistics of occurrence are shown in the \textbf{Number} column of \Cref{tab:stoptime}.
For RL-based LRM (QwQ), the model can not skip its thinking process in many cases.
We observe that better external thoughts (generated by a larger model such as Qwen-max) can effectively improve the chances of skipping thoughts.
Additionally, harder problems are more likely to induce rethinking.
Given CoT generated by Qwen-2.5-7b as an example, our experiments show that on the Math-500 dataset, the average difficulty level (identified by the `level' data of the dataset) of the problem with/without rethinking is 3.58/2.96 out of 5.

In contrast, distillation-based LRMs show a different behavior. 
These models skip the thinking process in all cases and start the final response when encountering the \texttt{</think>} token. 
This indicates that distillation-based LRMs may not truly ``understand'' the concept of reasoning or thinking.
Instead, their behavior is primarily driven by pattern-following skills learned during supervised fine-tuning~\cite{wulfmeier2024imitating, swamy2025all}.
Regarding the position where we place external thoughts, we try different approaches,e.g., inserting directly and within the chat template in supplementary materials.

\mypara{Hidden-State-Level Analysis}
To reinforce that the findings are linked to the model's inner perception rather than being a mere formatting issue, we analyze the generation dynamics of the \texttt{</think>} token from the logit perspective. 
Specifically, we compute the rank of the probability assigned to the \texttt{</think>} token at each decoding position. 
Let $V$ denote the vocabulary with size $|V|$, and let $\mathbf{z}_t \in \mathbb{R}^{|V|}$ be the logit vector at time step $t$. 
Let $\text{idx}_{\texttt{</think>}}$ be the index of the \texttt{</think>} token in the vocabulary. 
The rank of the \texttt{</think>} token at position $t$ is computed as:
\[
\text{Rank}_t = \left| \left\{ w' \in V \,\middle|\, \exp(z_t^{(w')}) > \exp(z_t^{(\text{idx}_{\texttt{</think>}})}) \right\} \right| + 1
\]
This rank reflects how strongly the model prioritizes emitting \texttt{</think>} at position t, with lower ranks indicating higher relative probability. 
We apply this computation across the full generation sequence and segment the input-output trace using special tokens: \texttt{<|im\_end|>} marks the end of the user input, and \texttt{</think>} marks the end of a reasoning segment. 
This allows us to identify and analyze three distinct phases: (1) the question phase, (2) the thinking phase (including external CoT and any rethinking), and (3) the final answer generation phase. 

The sample-level results are shown in~\Cref{fig:trajectory}, and the coarse-level results are shown in~\Cref{tab:stoptime}. 
We observe that during the thinking phase, the rank of \texttt{</think>} remains relatively high (i.e., low priority), indicating the model does not intend to terminate reasoning. 
As the model transitions toward the answer phase, the rank of \texttt{</think>} drops significantly, indicating a sharp increase in the likelihood of ending the thought process. 
These findings suggest the skip of thought is rooted in the model’s learned internal dynamics, not merely surface-level pattern matching. 

\begin{table}[ht]
\centering
\caption{\textbf{Efficacy Analysis}:
We report average reasoning tokens (\#Samples), steps (\#Steps), and double-checks (\#Checks) on MATH-500.  
Only cases with non-empty thoughts from the CoT generator are included.}
\label{tab:save}
\resizebox{\linewidth}{!}{
\begin{tabular}{c|c|c|c|c|c|c|c}
\toprule
\multicolumn{2}{c|}{\textbf{Ex.}} & \textbf{Skip?} & \textbf{\#Samples} & \textbf{Acc(\%)} & \textbf{\#Token} & \textbf{\#Steps} & \textbf{\#Check} \\
\midrule
\multirow{11}{*}{\rotatebox{90}{\textbf{QwQ}}}
& Baseline & -   & 500 & 88.47 & 4537 & 7.51 & 9.08 \\
\cmidrule{2-8}
& \multirow{2}{*}{3B} & Yes & 38  & 64.79 & 1775 & 4.84 & -    \\
&                    & No  & 48  & 89.51 & 3541 & 7.46 & 9.90 \\
\cmidrule{2-8}
& \multirow{2}{*}{7B} & Yes & 130 & 77.71 & 663  & 6.52 & -    \\
&                    & No  & 93  & 89.50 & 3269 & 7.32 & 8.31 \\
\cmidrule{2-8}
& \multirow{2}{*}{Plus} & Yes & 219 & 84.47 & 1212 & 6.17 & -    \\
&                       & No  & 281 & 89.68 & 4003 & 7.53 & 6.94 \\
\cmidrule{2-8}
& \multirow{2}{*}{Max} & Yes & 281 & 82.92 & 1041 & 6.39 & -    \\
&                      & No  & 219 & 87.67 & 4074 & 7.86 & 8.36 \\
\midrule
\multirow{7}{*}{\rotatebox{90}{\textbf{DPSK-14B}}}
& Baseline & -   & 500 & 65.80 & 2458 & 5.45 & 4.46 \\
\cmidrule{2-8}
& 3B       & Yes & 86  & 63.33 & 1052 & 6.19 & -    \\
\cmidrule{2-8}
& 7B       & Yes & 223 & 64.08 & 715  & 7.44 & -    \\
\cmidrule{2-8}
& Plus     & Yes & 500 & 67.20 & 595  & 7.39 & -    \\
\cmidrule{2-8}
& Max      & Yes & 500 & 66.06 & 635  & 7.35 & -    \\
\bottomrule
\end{tabular}
}
\end{table}

\subsection{Efficacy of External Thoughts}
Given the feasibility of skipping thinking, we conduct a fine-grained analysis to show whether LRMs can benefit from skipping the thoughts, especially on the efficiency aspect.

\mypara{Efficacy Metrics}
Based on the previous findings, we first categorize the questions based on whether the thinking process is skipped. 
We evaluate accuracy and token consumption to assess overall performance. 
Additionally, we count the number of reasoning steps and identify double-check conditions (determined by special tokens like ``Hmm'', ``Wait'', and ``Alternatively'').
Detailed calculations of these metrics are shown in the supplementary materials.

\mypara{Efficacy Analysis}
The results for QwQ and Deepseek-Distillation-Qwen-2.5-14B-Instruct on MATH-500 are presented in \Cref{tab:save}. 
We focus only on cases where the thought generator provides non-empty reasoning, i.e., it does not output \texttt{<STOP>}, which explains why the number of samples for smaller generators (e.g., 3b and 7b) is less than 500. 

We observe significant token reduction when internal thinking is skipped. 
In such cases, both RL-based and distillation-based LRMs proceed directly to answer generation, bypassing redundant reasoning steps and double-checking behaviors (as reflected in the \# Steps and \# Check columns). 
When the model does not skip thinking, token savings are still present but smaller, as the model generates additional internal reasoning. 

On average, accuracy in skipped cases is slightly lower than in non-skipped cases, indicating that some external CoTs may be unreliable or misaligned with the LRM’s reasoning style. 
However, this degradation is generally acceptable. 
For Plus and Max generators, accuracy drops by around 5\%, while token usage is reduced by over half. 
The main exception occurs with the 3b and 7b generators on the RL-based model, where performance degradation is more pronounced.
Yet, these cases are relatively few (38 and 130 out of 500), so their impact on overall performance remains limited. 
Further error analysis is provided in the experiment. 

Combining the identified pattern and the efficacy analysis, it is natural to develop an efficient inference pipeline for LRMs utilizing the external thoughts.

\begin{table*}[ht!]
\centering
\caption{\textbf{Overall Efficiency and Utility}: Accuracy (Acc) and token counts (Tokens) are reported. For ThoughtMani, the number in parentheses after Tokens indicates additional CoT tokens.}
\label{tab:main}
\renewcommand{\arraystretch}{0.90} 
\resizebox{0.9\textwidth}{!}{
\begin{tabular}{c|ccccccccccccccc}\toprule
 & \multicolumn{3}{c|}{AIME-2024} &  & \multicolumn{3}{c|}{GSM-8k} &  & \multicolumn{3}{c|}{MATH-500} &  & \multicolumn{3}{c}{Livebench/Coding} \\ \cmidrule{2-4} \cmidrule{6-8} \cmidrule{10-12} \cmidrule{14-16} 
\multirow{-2}{*}{Method} & Acc & Tokens & \multicolumn{1}{c|}{CoT} &  & Acc & Tokens & \multicolumn{1}{c|}{CoT} &  & Acc & Tokens & \multicolumn{1}{c|}{CoT} &  & { { Pass@k}} & Tokens & CoT \\ \midrule
 & \multicolumn{15}{c}{QwQ-32B} \\ \midrule
Full & 70.0 & 13661 & \multicolumn{1}{c|}{} &  & 95.3 & 1791 & \multicolumn{1}{c|}{} &  & 88.5 & 4537 & \multicolumn{1}{c|}{} &  & 66.7 & 6840 &  \\
Nothink & 40.0 & 12085 & \multicolumn{1}{c|}{} &  & 95.1 & 1552 & \multicolumn{1}{c|}{} &  & 80.4 & 4321 & \multicolumn{1}{c|}{} &  & 64.3 & 5865 &  \\
Prompt & 43.3 & 10897 & \multicolumn{1}{c|}{} &  & 93.1 & 665 & \multicolumn{1}{c|}{} &  & 82.2 & 3190 & \multicolumn{1}{c|}{} &  & 63.5 & 6518 &  \\
Truncation & 36.7 & 12508 & \multicolumn{1}{c|}{} &  & 95.7 & 1624 & \multicolumn{1}{c|}{} &  & 81.0 & 4938 & \multicolumn{1}{c|}{} &  & 57.8	 & 4128 &  \\
TokenSkip & 50.0 & 11172 & \multicolumn{1}{c|}{} &  & 94.4 & 536 & \multicolumn{1}{c|}{} &  & 86.8 & 3225 & \multicolumn{1}{c|}{} &  & 65.9 & 4269 &  \\
CoT-Valve & 74.4 & 14199 & \multicolumn{1}{c|}{\multirow{-6}{*}{-}} &  & 95.5 & 1697 & \multicolumn{1}{c|}{\multirow{-6}{*}{-}} &  & 89.2 & 4546 & \multicolumn{1}{c|}{\multirow{-6}{*}{-}} &  & 74.6 & 6714 & \multirow{-6}{*}{-} \\ \midrule
Ours - 3b & 70.0 & 14329 & \multicolumn{1}{c|}{11} &  & 95.3 & 1725 & \multicolumn{1}{c|}{7} &  & 86.1 & 4077 & \multicolumn{1}{c|}{22} &  & 65.6 & 6842 & 2 \\
Ours - 7b & 70.0 & 13101 & \multicolumn{1}{c|}{77} &  & 94.0 & 1075 & \multicolumn{1}{c|}{52} &  & 86.0 & 3526 & \multicolumn{1}{c|}{56} &  & 62.2 & 4409 & 120 \\
Ours - Plus & 75.6 & 11400 & \multicolumn{1}{c|}{209} &  & 93.5 & 961 & \multicolumn{1}{c|}{79} &  & 86.7 & 2792 & \multicolumn{1}{c|}{141} &  & 64.1 & 4461 & 137 \\
Ours - Max & 60.0 & 9607 & \multicolumn{1}{c|}{568} &  & 93.9 & 759 & \multicolumn{1}{c|}{132} &  & 85.6 & 2335 & \multicolumn{1}{c|}{209} &  & 60.9 & 4209 & 183 \\ \midrule
 & \multicolumn{15}{c}{Deepseek-Distillation-Qwen-2.5-32B-Instruct} \\ \midrule
Full & 68.9 & 9915 & \multicolumn{1}{c|}{} &  & 88.3 & 439 & \multicolumn{1}{c|}{} &  & 84.0 & 2973 & \multicolumn{1}{c|}{} &  & 60.2 & 6777 &  \\
Nothink & 43.3 & 9032 & \multicolumn{1}{c|}{} &  & 89.7 & 223 & \multicolumn{1}{c|}{} &  & 69.4 & 609 & \multicolumn{1}{c|}{} &  & 43.2 & 737 &  \\
Prompt & 50.0 & 8808 & \multicolumn{1}{c|}{} &  & 89.6 & 370 & \multicolumn{1}{c|}{} &  & 78.2 & 2167 & \multicolumn{1}{c|}{} &  & 57.3 & 5882 &  \\
Truncation & 30.0 & 4638 & \multicolumn{1}{c|}{} &  & 88.8 & 267 & \multicolumn{1}{c|}{} &  & 75.8 & 1760 & \multicolumn{1}{c|}{} &  & 54.7 & 10103 &  \\
TokenSkip & 40.0 & 3455 & \multicolumn{1}{c|}{} &  & 89.4 & 423 & \multicolumn{1}{c|}{} &  & 76.6 & 1567 & \multicolumn{1}{c|}{\multirow{-5}{*}{-}} &  & 49.5 & 6084 &  \\
CoT-Valve & 63.3 & 10359 & \multicolumn{1}{c|}{\multirow{-6}{*}{-}} &  & 88.8 & 478 & \multicolumn{1}{c|}{\multirow{-6}{*}{-}} &  & 82.1 & 2856 & \multicolumn{1}{c|}{} &  & 60.2 & 6012 & \multirow{-6}{*}{-} \\ \midrule
Ours - 3b & 62.2 & 10210 & \multicolumn{1}{c|}{11} &  & 88.3 & 415 & \multicolumn{1}{c|}{7} &  & 82.6 & 2526 & \multicolumn{1}{c|}{22} &  & 59.1 & 6557 & 2 \\
Ours - 7b & 54.3 & 7985 & \multicolumn{1}{c|}{77} &  & 86.8 & 292 & \multicolumn{1}{c|}{52} &  & 79.4 & 2170 & \multicolumn{1}{c|}{56} &  & 41.7 & 528 & 120 \\
Ours - Plus & 20.1 & 2076 & \multicolumn{1}{c|}{209} &  & 87.5 & 263 & \multicolumn{1}{c|}{79} &  & 68.3 & 554 & \multicolumn{1}{c|}{141} &  & 45.8 & 528 & 137 \\
Ours - Max & 21.1 & 1482 & \multicolumn{1}{c|}{568} &  & 88.7 & 267 & \multicolumn{1}{c|}{132} &  & 67.8 & 562 & \multicolumn{1}{c|}{209} &  & 44.5 & 465 & 183 \\ \midrule
 & \multicolumn{15}{c}{Deepseek-Distillation-Qwen-2.5-14B-Instruct} \\ \midrule
Full & 31.1 & 8273 & \multicolumn{1}{c|}{} &  & 87.6 & 756 & \multicolumn{1}{c|}{} &  & 65.3 & 2392 & \multicolumn{1}{c|}{} &  & 54.7 & 6871 &  \\
Nothink & 30.0 & 8215 & \multicolumn{1}{c|}{} &  & 75.2 & 216 & \multicolumn{1}{c|}{} &  & 63.8 & 796 & \multicolumn{1}{c|}{} &  & 33.6 & 657 &  \\
Prompt & 33.3 & 8803 & \multicolumn{1}{c|}{} &  & 88.3 & 516 & \multicolumn{1}{c|}{} &  & 65.2 & 1904 & \multicolumn{1}{c|}{} &  & 54.9 & 6312 &  \\
Truncation & 26.7 & 5204 & \multicolumn{1}{c|}{} &  & 84.2 & 214 & \multicolumn{1}{c|}{} &  & 62.6 & 1627 & \multicolumn{1}{c|}{} &  & 46.9 & 9245 &  \\
TokenSkip & 30.0 & 8503 & \multicolumn{1}{c|}{} &  & 89.3 & 314 & \multicolumn{1}{c|}{} &  & 73.2 & 1356 & \multicolumn{1}{c|}{} &  & 0.0 & 10750 &  \\
CoT-Valve & 15.0 & 10967 & \multicolumn{1}{c|}{\multirow{-6}{*}{-}} &  & 86.7 & 681 & \multicolumn{1}{c|}{\multirow{-6}{*}{-}} &  & 62.9 & 2190 & \multicolumn{1}{c|}{\multirow{-6}{*}{-}} &  & 56.2 & 6042 & \multirow{-6}{*}{-} \\ \midrule
Ours - 3b & 19.9 & 8649 & \multicolumn{1}{c|}{11} &  & 86.4 & 691 & \multicolumn{1}{c|}{7} &  & 65.2 & 2080 & \multicolumn{1}{c|}{22} &  & 53.9 & 6670 & 2 \\
Ours - 7b & 24.4 & 7952 & \multicolumn{1}{c|}{77} &  & 85.7 & 356 & \multicolumn{1}{c|}{52} &  & 69.2 & 1742 & \multicolumn{1}{c|}{56} &  & 38.5 & 588 & 120 \\
Ours - Plus & 16.6 & 2209 & \multicolumn{1}{c|}{209} &  & 88.1 & 272 & \multicolumn{1}{c|}{79} &  & 65.4 & 600 & \multicolumn{1}{c|}{141} &  & 39.6 & 625 & 137 \\
Ours - Max & 18.8 & 1838 & \multicolumn{1}{c|}{568} &  & 89.6 & 281 & \multicolumn{1}{c|}{132} &  & 64.6 & 595 & \multicolumn{1}{c|}{209} &  & 37.0 & 523 & 183\\\bottomrule
\end{tabular}}
\end{table*}

\begin{algorithm}[t!]
\caption{\Method Pipeline}
\label{alg:cot_reasoning}
\KwIn{A dataset $D = \{q_1, q_2, \dots, q_n\}$ containing problems, a CoT generator model $G$, a reasoning model $M$}
\KwOut{Final responses $\{r_1, r_2, \dots, r_n\}$ for each problem in $D$.}

$\mathcal{T}_{\text{Mani}} \gets \texttt{\textless|im\_start|\textgreater} ~ {User:} ~ \textbf{[Question]} ~ \texttt{\textless|im\_end|\textgreater}  \newline
\texttt{\textless|im\_start|\textgreater} ~ {Assistant:} ~ \texttt{\textless|im\_end|\textgreater}  \newline
\texttt{\textless think\textgreater} ~ \textbf{[CoT]} ~ \texttt{\textless /think\textgreater}$ \;

$\mathcal{T}_{\text{Ori}} \gets \texttt{\textless|im\_start|\textgreater} ~ {User:} ~ \textbf{[Question]} ~ \texttt{\textless|im\_end|\textgreater}  \newline
\texttt{\textless|im\_start|\textgreater} ~ {Assistant:} ~ \texttt{\textless|im\_end|\textgreater}  \newline
\texttt{\textless think\textgreater}$ \;

\For{each problem $q_i \in D$}{
    $C_i \gets G(q_i)$ \tcp{Generate CoTs}
    
    \If{$C_i = \texttt{<STOP>}$}{
        $T_i \gets \mathcal{T}_{\text{Ori}}[\text{Question} \gets q_i]$ \tcp{Format $\mathcal{T}_{\text{Ori}}$ with $q_i$}
    }
    \Else{
        $T_i \gets \mathcal{T}_{\text{Mani}}[\text{Question} \gets q_i, \text{CoT} \gets C_i]$ \tcp{Format $\mathcal{T}_{\text{Mani}}$ with $q_i$ and $C_i$}
    }
    
    $r_i \gets M(T_i)$ \tcp{Obtain final response from  $M$}
    
    Append $r_i$ to the output set $\{r_1, r_2, \dots, r_n\}$ \;
}

\Return $\{r_1, r_2, \dots, r_n\}$ 
\end{algorithm}

\subsection{Pipeline of \Method}
\label{sec:method}

Based on these observations, we further propose our method, \Method, to leverage these behaviors and improve reasoning efficiency in LRMs.
Specifically, when the CoT generator outputs only the placeholder \texttt{<STOP>}, indicating that it cannot provide useful reasoning, we discard the empty CoT and fall back to the original inference template. 
This ensures that the LRM retains full control over the reasoning process for challenging problems, avoiding potential degradation from incorrect or misleading external thoughts.
The detailed pipeline is shown in \Cref{alg:cot_reasoning}.

\section{Experiment}

\subsection{Experimental Setup}
\mypara{Datasets}
To evaluate the effectiveness of the proposed CoT-reduced reasoning process, we select four different datasets, covering reasoning ability in both math and coding.
For reasoning, we select three math datasets, including AIME-2024~\cite{maxwelljia_aime2024}, GSM-8k~\cite{cobbe2021gsm8k} and MATH-500~\cite{lightman2023lets}.
For coding, we select the coding category from LiveBench~\cite{white2024livebench}.

\mypara{Metrics}
We quantify the performance from three perspectives, i.e., utility, efficiency.
For utility, we extract answers via string matching for the AIME, GSM-8k, and MATH-500 datasets. 
Regarding the coding dataset, we follow the official guidance and report the pass@1 metric on private test cases. 
For efficiency, we compute the generated tokens from the reasoning model and the additional tokens produced by the CoT generators.
The total cost of generation is evaluated as the sum of these two components. 
Since the CoTs are generated by smaller models, the cost of producing these additional tokens is significantly lower compared to the computational expense of the larger reasoning model.

\mypara{Models}
Regarding CoT generators, we consider Qwen-series~\cite{yang2024qwen2}, including Qwen-Max, Qwen-Plus, Qwen-2.5-7B-Instruct, and Qwen-2.5-3B-Instruct.
Since we need to accurately manipulate the inference template in the stated approach, we only consider local open-source LRMs.
Regarding RL-based LRMs, we select QwQ-32B~\cite{qwq32b}, which is derived by RL from Qwen-2.5-32B-Instruct.
Regarding distillation-based LRMs, we select Deepseek-Distillation-Qwen-2.5-14B-Instruct and its 32B version, which distills the CoT generated from DeepSeek-R1 on Qwen series~\cite{DBLP:journals/corr/abs-2501-12948}.
More settings are shown in the supplementary materials.

\mypara{Baselines}
We take the following methods as our baselines, and more details are in the supplementary materials: 
\begin{itemize}[leftmargin=*,itemsep=0.1em, topsep=0.2em]
\item \textit{Nothink}~\cite{ma2025reasoning} places ``I have finished the thoughts'' within the thinking tokens, as \Method does.

\item \textit{Truncation} directly terminates the thinking process by interrupting the generation when a predefined thinking budget is met and inserting a \texttt{</think>} token to output the answer.
Specifically, we cut 50\% of the length of the original thinking process (full thinking by the original model).

\item \textit{Prompt Reduction}~\cite{ding2024break} provides instructions like ``Let’s quickly conclude the answer without showing step-by-step reasoning.'' to reduce the thinking process.
\item{\textit{Tokenskip}~\cite{xia2025tokenskip}} first constructs a dataset where less important tokens are pruned and fine-tunes the models on the compressed dataset to enable the model to selectively skip the redundant tokens in inference. 

\item \textit{CoT-Valve}~\cite{ma2025cot} utilizes interpolation of the LLMs' and LRMs' parameters to collect CoTs of varying lengths, followed by progressively fine-tuning the LRMs to compress the length of CoT.
\end{itemize}
\begin{figure}[t]
    \centering
    \includegraphics[width=\linewidth]{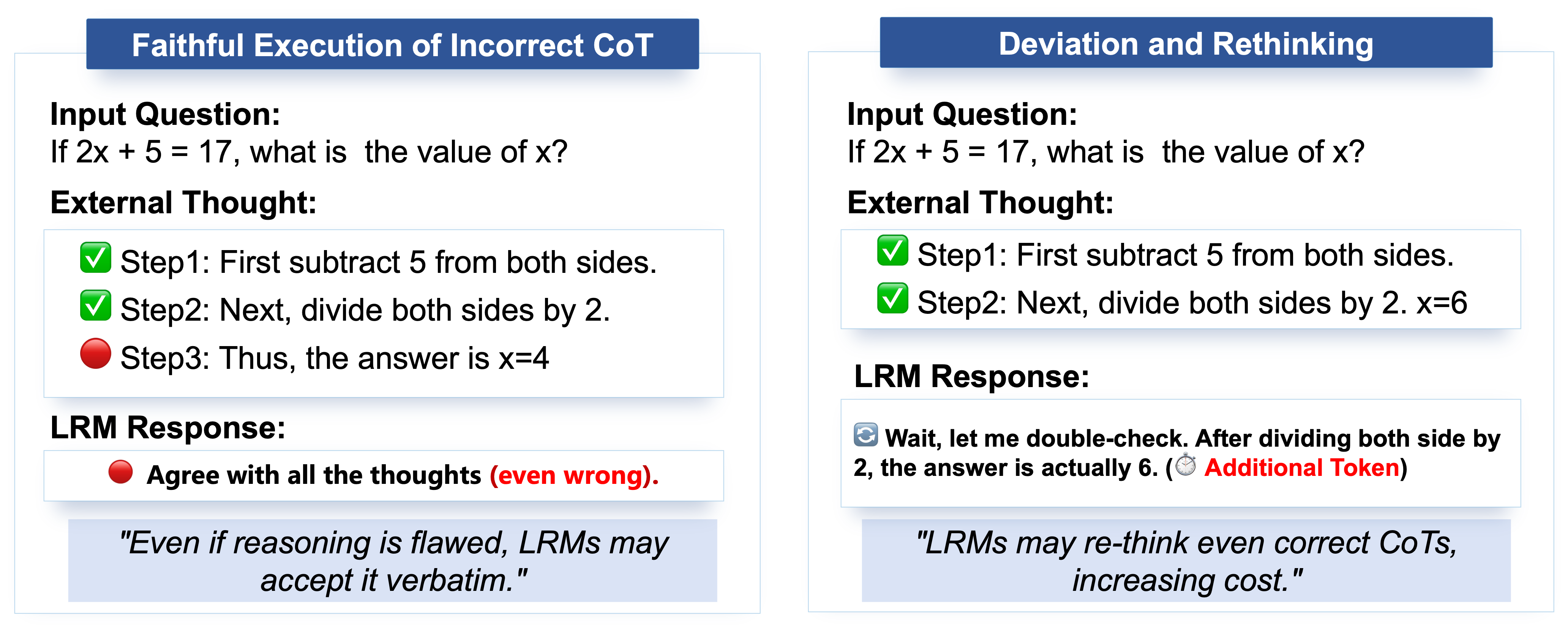}
    \caption{
    \textbf{Two Suboptimal Modes}: 
(Left) The LRM faithfully follows an incorrect external CoT, resulting in a wrong answer. 
(Right) The LRM rejects the external reasoning and generates additional thoughts, leading to high token consumption despite eventual correctness. 
}
    \label{fig:error}
\end{figure}

\begin{figure}[t]
    \centering
    \includegraphics[width=1\linewidth]{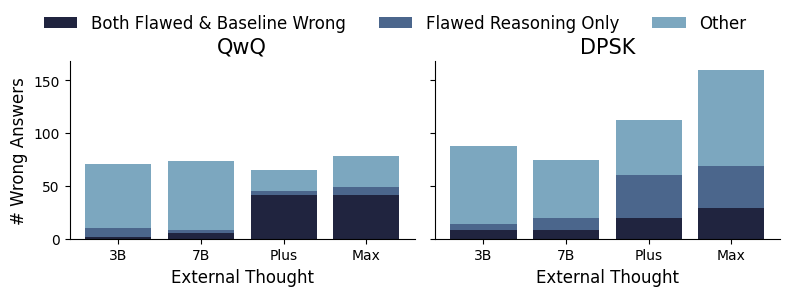}
    \includegraphics[width=1\linewidth]{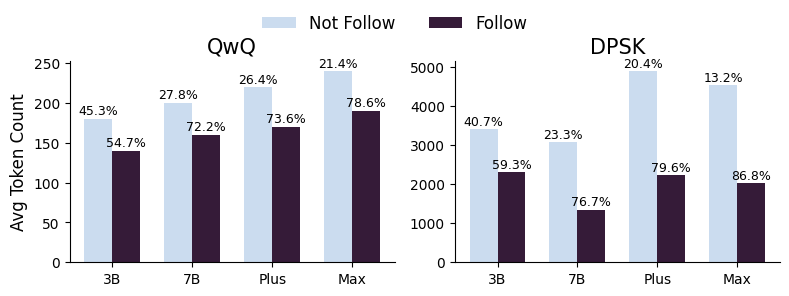}

    \caption{\textbf{Suboptimal Mode Analysis} (Upper) Visualization of how flawed reasoning impacts model performance, illustrating cases of degradation (baseline correct but misled by flawed thought), consistently wrong predictions (baseline incorrect despite flawed reasoning), and correct reasoning with incorrect final answers. (Lower) Bar plots depict average token counts for outputs that adhere to versus deviate from the provided reasoning, with numerical values indicating the percentage occurrence of each case.}
    \label{fig:error-a}
\end{figure}

\subsection{Overall Efficiency and Utility}
The main results of our experiments are shown in \Cref{tab:main}.
\textit{Nothink} can effectively reduce the tokens on the distillation-based models at the cost of performance, while showing limited effects on RL-based models.
\textit{Prompt Reduction} and \textit{Truncation} can decrease token counts to some extent, but the reduction varies unpredictably, and the associated performance drop can be substantial.
For \textit{Tokenskip}, the performance of in-domain cases, i.e., GSM-8k and MATH-500, is competitive in both utility and efficiency, while showing limited ability to transfer to other datasets. 
For \textit{CoT-Vavle}, the reproduced performance shows increased utility while the compression ability is usually. 

Generally, \Method shows competitive performance.
For the RL-based model (QwQ), \Method with four different CoT generators reduces the response length by 1\%, 18\%, 26\%, and 37\% with 1.5\%, 2.8\%, 0.8\%, and 7.2\% performance drop for the average on four different datasets.
For the distillation-based models, \Method with four different CoT generators reduces the response length by 2\%, 45\%, 82\%, and 86\% with a relatively higher 4.5\%, 11.5\%, 20.4\%, and 18.2\% performance drop for the average on four different datasets.
Since smaller CoT generators may refuse to provide CoT in many hard cases and return empty thoughts, this makes the average CoT length relatively short.

\subsection{Suboptimal Mode Analysis}
We analyze the outputs of \Method and identify two primary suboptimal modes that affect accuracy and efficiency. 
First, when the external thought contains flawed reasoning, such as an incorrect final inference despite mostly sound logic, the LRM may blindly follow this erroneous path, resulting in a wrong final answer (\Cref{fig:error}, left). 
Second, the LRM may reject the external reasoning, generating new thoughts or performing redundant checks (e.g., “Hmm”, “Wait”), which increases token usage even when the final answer is correct (\Cref{fig:error}, right). 
To systematically identify these error types, we prompt a strong LLM to judge whether the generated response contains reasoning flaws or diverges from the external CoT. 
The prompts and representative cases are included in the supplement. 

On the wrongly answered subset of MATH-500 (\Cref{fig:error-a}, Upper), we further analyze the first suboptimal type by splitting it into two subcases: whether the problem can be solved correctly under vanilla inference (without external thoughts). 
When the original model also fails on a problem, the impact of flawed external reasoning is limited, since performance is already near the ceiling. 
The more harmful case occurs when a problem that could have been solved correctly under standard inference becomes erroneous due to misleading external thoughts. 
This degradation is particularly evident in distillation-based models, which lack the ability to re-think and are thus more vulnerable to faulty CoTs. 

For the second suboptimal type (\Cref{fig:error-a}, Lower), divergence from external reasoning, token consumption increases due to additional internal reasoning steps. 
While this reduces efficiency, the ``unfollow'' behavior often helps the model correct flawed or incomplete external thoughts, thereby improving robustness and accuracy. 
As such, this mode is less harmful overall and can even be beneficial in maintaining performance when external CoTs are unreliable. 

\mypara{Difficulty-Aware Mitigation}
We focus on the harmful scenario in the first error mode—where flawed external reasoning leads the model to make mistakes on problems it could otherwise solve correctly.
This issue is particularly pronounced on difficult queries, where larger CoT generators attempt to reason but produce incorrect or misleading steps. 
As a simple mitigation, we exclude external thoughts for level-5 problems (the most challenging) in the MATH-500 dataset and allow the LRM to generate reasoning from scratch.

We apply this adjustment specifically to distillation-based models paired with larger CoT generators (e.g., Plus and Max), which are most vulnerable to this failure mode due to their inability to re-think. 
As shown in \Cref{tab:mitigate}, this approach significantly improves accuracy and reduces the harmful cases, while still achieving substantial token savings compared to the baseline. 
The results suggest that a practical enhancement to \Method is to reserve internal reasoning for the hardest problems, effectively balancing efficiency and reliability. 
Although recent work has investigated difficulty-aware and adaptive reasoning mechanisms~\cite{liu2025think, zhu2025towards}, integrating such methods is left for future exploration. 

\begin{table}[]
\centering
\caption{\textbf{Error Mitigation}: We report the accuracy and token consumption using the mitigation. \# degrades represents the number of degradation cases.}
\label{tab:mitigate}
\resizebox{\linewidth}{!}{

\begin{tabular}{c|c|ccc|ccc}\toprule
\multirow{2}{*}{Ex.} & \multirow{2}{*}{Method} & \multicolumn{3}{c|}{DPSK-14B} & \multicolumn{3}{c}{DPSK-32B} \\ \cmidrule{3-8} 
 &  & Acc & Token & \# degrade & Acc & Token & \# degrade \\ \midrule
- & Baseline & 65.3 & 2392 & - & 84 & 2973 & - \\ \midrule
\multirow{2}{*}{Plus} & Ours & 65.4 & 600(141) &  37& 68.3 & 554(141) & 43 \\
 & Enhanced & 66.4\textsubscript{{↑1.0}} & 1381(96) &  20& 75.0\textsubscript{{↑6.7}} & 1752(96) & 18 \\ \midrule
\multirow{2}{*}{Max} & Ours & 64.6 & 595(209) &  34& 67.8 & 562(209) & 40 \\
 & Enhanced & 69.4\textsubscript{{↑4.8}} & 1398(143) &  15& 75.8\textsubscript{{↑8.0}} & 1684(143) & 19\\\bottomrule
\end{tabular}
}\end{table}

\subsection{Execution Time}
As shown in \Cref{tab:time}, we measure the execution time of \Method on the MATH-500 dataset to evaluate the real-world inference speedup. 
For distillation-based LRMs, the execution time is reduced by approximately 25\%, while for RL-based models, the improvement reaches nearly 50\%. 
This significant reduction in latency demonstrates that \Method not only saves tokens but also delivers tangible acceleration in practice, highlighting its effectiveness and practical value for real-world applications where efficiency and response speed are critical.

\subsection{Other Evaluation}
We further study the generalization ability of our approach across different tasks and the impact of token budgets on performance. 
Regarding the generalization ability, we utilize task-specific prompts to generate the external thoughts on the Code dataset, but observe no significant performance difference between the two settings. 
Regarding the token budget, we limit the token budget of generation and find that the token reduction persists at all budget levels.
Due to length limitations, we leave the detailed experimental settings and results for the supplementary materials.

\section{Related Work}
\mypara{Large Reasoning Model}
By scaling up training data size and model size, large language models (LLMs) have developed powerful language understanding and generation capabilities~\cite{DBLP:journals/corr/abs-2303-18223}, such as GPT-4o~\cite{DBLP:journals/corr/abs-2410-21276} and DeepSeekv3~\cite{DBLP:journals/corr/abs-2412-19437}, enabling rapid and coherent responses to user inputs.
However, these models perform poorly when facing complex reasoning and logical analysis tasks~\cite{xu2025towards,DBLP:journals/corr/abs-2303-18223}, falling far short of human cognitive levels. 
To address this issue, recent studies focus on improving the capability of language models by utilizing more inference-time computation instead of simply scaling model parameters~\cite{snell2024scaling}.
This line of research has already outcome many powerful LRMs such as DeepSeek-R1~\cite{DBLP:journals/corr/abs-2501-12948}, OpenAI-o1/o3~\cite{openai2025o1,openai2025o3mini}, and QwQ~\cite{qwq32b}, which shift from fast, intuitive processing to structured, step-by-step reasoning. 
Deeper reasoning capabilities enable these LRMs to make remarkable improvements when tackling challenging tasks like advanced mathematics and logical reasoning~\cite{DBLP:journals/corr/abs-2502-14739}.
\begin{table}[t]
\centering
\caption{\textbf{Execution Time}: we record the execution time of running \Method and baseline on the MATH-500 dataset. The reported number is the average running time.}
\label{tab:time}
\resizebox{\linewidth}{!}{\begin{tabular}{c|c|ccc|c}
\toprule
\multirow{2}{*}{Model} & \multirow{2}{*}{Baseline} & \multicolumn{3}{c|}{\Method} & \multirow{2}{*}{Improve} \\ \cmidrule{3-5}
 &  & Thoughts & Inference & Total &  \\ \midrule
QwQ & 19.29 & \multirow{3}{*}{0.20} & 9.87 & 10.07 & 47.80\% \\
DPSK-32B & 10.11 &  & 7.35 & 7.55 & 25.32\% \\
DPSK-14B & 3.41 &  & 2.41 & 2.61 & 23.46\% \\\bottomrule
\end{tabular}
}\end{table}

\mypara{Chain-of-Thought Compression}
Despite the improved capabilities, introducing intermediate CoTs brings additional token overhead.
For traditional LLMs, lots of efforts have been put into reducing redundant steps~\cite{ding2024break,liu2024can,ma2025cot} or skipping less-important tokens~\cite{han2024token, xia2025tokenskip}
Another line is to represent the CoT using latent space compression~\cite{deng2024explicit,cheng2024compressed}, which utilizes embeddings instead of tokens to serve as the CoT.
With the development of LRMs, many works focus on solving the overthinking problem~\cite{chen2024not, fan2025missing, zhu2025towards,sui2025stop}.

Our work is similar to speculative decoding~\cite{leviathan2023fast}, but differs in that we use a smaller model only to generate reasoning thoughts, enabling the main model to skip internal thinking and reduce token generation. 
Speculative decoding, in contrast, uses a small model to guess output tokens without shortening the reasoning process. 
Another similar method is NoThink~\cite{ma2025reasoning}, which inserts a placeholder sentence to avoid thinking. 
While similar in spirit, our approach goes further by using meaningful external thoughts and revealing how RL-based and distillation-based models differ in their response to external reasoning—enabling more informed and effective reasoning skipping.

\section{Conclusion}

In this paper, we propose \Method, a simple yet efficient inference pipeline, and reveal an important characteristic of LRM behaviors.
Through extensive experiments, we demonstrate that \Method can significantly reduce computational costs while keeping the utility.
By providing a practical solution to improve efficiency without compromising utility, \Method makes LRMs more accessible for real-world, resource-constrained applications.

\mypara{Implications}
Our findings highlight significant implications for LRMs. 
By understanding the distinct behaviors of RL-trained and distillation-trained LRMs, we can make better-informed deployment decisions. 
Since model vendors typically serve models of different sizes simultaneously, \Method offers a practical solution to reduce computational costs while maintaining accuracy, making LRMs more efficient and accessible for real-world scenarios.
\bibliographystyle{plain}
\bibliography{ref}
\appendix
\newpage
\section{Supplement for Effectiveness of External Thoughts}
\subsection{Sample Output of Replacing Thoughts}
\begin{cotbox}[Thinking is Skipped]
    \texttt{<|im\_start|>} 
    \textbf{\color{maincolor} User:} 
    If $f(x)=\dfrac{2}{x+1}$, then what is the value of $f^{-1}\left(\frac{1}{5}\right)$?
    \texttt{<|im\_end|>}

    \texttt{<|im\_start|>} 
    \textbf{\color{maincolor} Assistant:} 
    \texttt{<|im\_end|>}

    \texttt{<think>}
    
    \textbf{\color{highlight} [External Thoughts]}
    
    \textbf{ Understand the inverse function:} 
    The inverse function $ f^{-1}(y) $ is defined such that if $ y = f(x) $, then $ x = f^{-1}(y) $.

    \textbf{ Set up the equation:} 
    Set $ y = \frac{1}{5} $ and solve $ \frac{1}{5} = \frac{2}{x+1} $.

    \textbf{ Solve for $ x $:} 
    Cross-multiplying gives $ 1 \cdot (x + 1) = 5 \cdot 2 $, so $ x + 1 = 10 $, hence $ x = 9 $.  
    Therefore, $ f^{-1}\left(\frac{1}{5}\right) = 9 $.
    \texttt{</think>}
    \tcblower
    \textbf{\color{highlight} [Final Response]}
        To find \( f^{-1}\left(\frac{1}{5}\right) \) for the function \( f(x) = \dfrac{2}{x+1} \):

    1. **Set the function equal to \( \frac{1}{5} \)**:
       \[
       \frac{2}{x+1} = \frac{1}{5}
       \]

    2. **Solve for \( x \)**:
       - Cross-multiply to eliminate the denominators:
         \[
         2 \cdot 5 = 1 \cdot (x + 1)
         \]
       - Simplify:
         \[
         10 = x + 1
         \]
       - Subtract 1 from both sides:
         \[
         x = 10 - 1 = 9
         \]

    Thus, the value of \( f^{-1}\left(\frac{1}{5}\right) \) is \( \boxed{9} \).

\end{cotbox}

\begin{cotbox}[Thinking is Not Skipped (Rethink)]
    \texttt{<|im\_start|>} 
    \textbf{\color{maincolor} User:} 
    A regular hexagon can be divided into six equilateral triangles. If the perimeter of one of the triangles is 21 inches, what is the perimeter, in inches, of the regular hexagon?
    \texttt{<|im\_end|>}

    \texttt{<|im\_start|>} 
    \textbf{\color{maincolor} Assistant:} 
    \texttt{<|im\_end|>}

    \texttt{<think>} 

    \textbf{\color{highlight} [External Thoughts]}
    
    To solve this problem, consider the following key points:
    \begin{itemize}
        \item A regular hexagon can be divided into six equilateral triangles.
        \item The perimeter of one of these triangles is given as 21 inches.
        \item Since the triangle is equilateral, all three sides are equal. Thus, each side of the triangle is 7 inches (21 inches / 3).
        \item Each side of the equilateral triangle is also a side of the hexagon.
        \item A regular hexagon has six equal sides.
        \item Therefore, the perimeter of the hexagon is 6 times the length of one side of the triangle.
    \end{itemize}
    \texttt{</think>}
            \tcblower

    \textbf{\color{highlight} [Rethinking]} Okay, let me try to figure this out. So, the problem is about a regular hexagon divided into six equilateral triangles. The perimeter of one of those triangles is 21 inches, and I need to find the perimeter of the hexagon. Hmm, let's start breaking it down step by step.

    First, I know that a regular hexagon has all sides equal, and when it's divided into six equilateral triangles, those triangles must be using the sides of the hexagon as their sides. So, each triangle is equilateral, meaning all three sides are the same length.

    The problem says the perimeter of one of these triangles is 21 inches. Since the triangle is equilateral, all three sides are equal. To find the length of one side, I can divide the perimeter by 3. Let me write that down:

    Perimeter of triangle = 21 inches \\
    Each side of the triangle = $ 21 \div 3 = 7 $ inches

    Okay, so each side of the triangle is 7 inches. Now, since the hexagon is regular, all its sides are equal too. And since the triangles are formed by dividing the hexagon, each side of the triangle must correspond to a side of the hexagon. Wait, actually, let me make sure that's the case.

    If you imagine a regular hexagon, and then draw lines from the center to each vertex, that divides it into six equilateral triangles. Each of those triangles has two sides that are radii of the hexagon and one side that is a side of the hexagon. Wait a second, hold on. Maybe I need to clarify the structure here.

    In a regular hexagon, when you connect the center to all six vertices, each of those six triangles is indeed equilateral. Each triangle has two sides that are equal to the radius of the circumscribed circle (the distance from the center to a vertex), and the third side is a side of the hexagon. But in an equilateral triangle, all sides must be equal. So, that would mean that the radius (the distance from the center to a vertex) is equal to the length of the side of the hexagon.

    Hmm, actually, that must be the case because if the triangle is equilateral, all three sides must be equal. Therefore, the two sides that are radii must be equal in length to the side of the hexagon. So, each of those triangles has sides equal to the side length of the hexagon. So, in that case, the perimeter of each triangle would be three times the side length of the hexagon.

    Therefore, if the perimeter of the triangle is 21 inches, each side of the triangle (and hence each side of the hexagon) is $ 21 \div 3 = 7 $ inches, as I calculated earlier.

    Since the hexagon has six sides, each 7 inches long, the perimeter of the hexagon would be $ 6 \times 7 = 42 $ inches.

    Wait, that seems straightforward, but let me just verify once more to be sure.

    Another way to think about this: The side of the equilateral triangle is the same as the side of the hexagon because the triangle is formed by connecting the center to two adjacent vertices. So the base of the triangle (the side that's part of the hexagon's perimeter) is equal in length to the other two sides (the radii). Therefore, each side of the triangle is equal to the side of the hexagon.

    Given that, the perimeter of the triangle is $ 3s = 21 $, so solving $ 3s = 21 $ gives $ s = 7 $. Then the hexagon's perimeter is $ 6s = 6 \times 7 = 42 $.

    Hmm, seems consistent. I don't see any errors in that logic. Let me just visualize it again. Imagine a regular hexagon divided into six triangles with the center. Each triangle is equilateral, so all their sides are equal. Therefore, each triangle's sides must be equal to the edges of the hexagon. So, the perimeters relate as described.

    Therefore, I think the correct answer is 42 inches.

    \textbf{Final Answer} \\
    The perimeter of the regular hexagon is $ \boxed{42} $ inches.
    \texttt{</think>} 
    
\textbf{\color{highlight} [Final Response]} A regular hexagon can be divided into six equilateral triangles, each with a perimeter of 21 inches. To find the perimeter of the hexagon:

1. \textbf{Determine the side length of the equilateral triangle:}
   - The perimeter of one equilateral triangle is 21 inches.
   - Since all sides of an equilateral triangle are equal, each side length is \( \frac{21}{3} = 7 \) inches.

2. \textbf{Relate the side length of the triangle to the hexagon:}
   - Each side of the equilateral triangle corresponds to a side of the hexagon.
   - Therefore, each side of the hexagon is 7 inches.

3. \textbf{Calculate the perimeter of the hexagon:}
   - A regular hexagon has six equal sides.
   - The perimeter is \( 6 \times 7 = 42 \) inches.

Thus, the perimeter of the regular hexagon is \( \boxed{42} \) inches.

\end{cotbox}
\mypara{Code Example for \Method Template}
\begin{lstlisting}[language=Python]
def process_example(question, external_thought):
    messages = [
        {"role": "user", "content": f"Question: {question}"},
        {"role": "assistant", "content": ""}
    ]
    template = tokenizer.apply_chat_template(
        messages,
        add_generation_prompt=True,
        tokenize=False
    )
    if not template.endswith(('<think>','<think>\n')):
        template += '</think>'
    template +=  f'{external_thought}\n\n</think>'
    return template
\end{lstlisting}

\subsection{Token-Level Analysis: Optimal Position}
Additionally, we conduct an ablation study to show the effectiveness of placing CoTs, enclosed by \texttt{<think>} and \texttt{</think>}, at the end of the chat template.
Specifically, we design the following two cases and represent them using chat templates, where the dashed line represents the start position for model generation.

For the first case, we consider placing the CoTs enclosed by thinking tokens within the template, i.e., before \texttt{<|im\_end|>}.
The experiment results are shown in \Cref{tab:inchat}.

\begin{cotbox}[Inference Template - Case 1 (within chat)]
    \texttt{<|im\_start|>} 
    \textbf{\color{maincolor} User:} \textbf{[Question]}
    \texttt{<|im\_end|>}

    \texttt{<|im\_start|>} 
    \textbf{\color{maincolor} Assistant:}\newline
    \texttt{<think>} \newline
    \textbf{\color{highlight} [Generated Thought]}\newline
    \texttt{</think>}\newline
    \texttt{<|im\_end|>}
    \tcblower

\end{cotbox}
\begin{table}[h]
\centering
\caption{Number of occurrences of \texttt{</think>} in the response using different models to generate CoT, which indicates the frequency of rethinking. The dataset is sampled from the original one.}
\label{tab:inchat}
\resizebox{\linewidth}{!}{\begin{tabular}{c|c|cccc}
\toprule
\multirow{2}{*}{\textbf{Dataset}} & \multirow{2}{*}{\textbf{Model}} & \multicolumn{4}{c}{\textbf{CoT Template -- within chat}} \\
 &  & 3b & 7b & plus & max \\ \midrule
\multirow{3}{*}{GSM-8k} & Distill-qwen-14b & 86 & 88 & 85 & 86 \\
 & Distill-qwen-32b & 0 & 0 & 0 & 0 \\
 & QwQ-32B & 98 & 97 & 86 & 89 \\ \midrule
\multirow{3}{*}{MATH-500} & Distill-qwen-14b & 164 & 160 & 172 & 155 \\
 & Distill-qwen-32b & 9 & 1 & 1 & 0 \\
 & QwQ-32B & 193 & 183 & 184 & 163
 \\\bottomrule
\end{tabular}}
\end{table}
For the second case, we consider directly placing the CoTs enclosed after the template, i.e., after \texttt{<|im\_end|>} but without thinking tokens.
The experiment results are shown in \Cref{tab:nothinking}.

\begin{cotbox}[Inference Template - Case 2 (no thinking token)]
    \texttt{<|im\_start|>} 
    \textbf{\color{maincolor} User:} \textbf{[Question]}
    \texttt{<|im\_end|>}

    \texttt{<|im\_start|>} 
    \textbf{\color{maincolor} Assistant:} 
    \texttt{<|im\_end|>}\newline
    \textbf{\color{highlight} [Generated Thought]}
        \tcblower

\end{cotbox}

\begin{table}[h]
\centering
\caption{Number of occurrences of \texttt{</think>} in the response using different models to generate CoT, which indicates the frequency of rethinking. The dataset is sampled from the original one.}
\label{tab:nothinking}
\resizebox{\linewidth}{!}{\begin{tabular}{c|c|cccc}
\toprule

\multirow{2}{*}{\textbf{Dataset}} & \multirow{2}{*}{\textbf{Model}} & \multicolumn{4}{c}{\textbf{CoT Template -- no thinking}} \\
 &  & 3b & 7b & plus & max \\ \midrule
\multirow{3}{*}{GSM-8k} & Distill-qwen-14b & 86 & 87 & 91 & 81 \\
 & Distill-qwen-32b & 0 & 0 & 0 & 0 \\
 & QwQ-32B & 97 & 81 & 83 & 55 \\ \midrule
\multirow{3}{*}{MATH-500} & Distill-qwen-14b & 165 & 161 & 172 & 163 \\
 & Distill-qwen-32b & 12 & 2 & 0 & 0 \\
 & QwQ-32B & 195 & 193 & 180 & 145
  \\\bottomrule

\end{tabular}}
\end{table}

\subsection{Metric of Efficacy}
\mypara{Double-check Condition} It measures how often the model shows signs of reconsidering or revising its reasoning before reaching a conclusion. 
This is detected by looking for specific words such as "Hmm", "Wait", and "Alternatively" in the part of the response that comes before the \texttt{</think>} tag. 
These words suggest the model is pausing, correcting itself, or exploring different ideas. 
The total count of such words is calculated using case-insensitive exact matching. 
If the </think> tag is missing, the count is zero. 
To avoid unrealistic values, any count of 30 or higher is treated as zero. 

\mypara{Reasoning Step} It reflects how the model breaks down its thinking into distinct parts after the </think> tag. 
The text is split into segments wherever a new step in reasoning begins, as identified by a regular expression that detects structural shifts such as line breaks or numbered points. 
The number of resulting segments is taken as the count of reasoning steps. 
If the text is empty or no boundaries are found, the entire content is treated as a single segment. 
As with double-checks, if the number of segments reaches 30 or more, it is set to zero to prevent distortion from unusually long or fragmented outputs. 

Together, these metrics help assess the model’s reasoning style—how carefully it checks its work and how clearly it structures its thoughts—offering insight beyond just the correctness of the final answer. 
\section{Supplement for Experiment}
\subsection{Experimental Settings}
\mypara{Model Setting}
Specifically, we use a 4-bit AWQ~\cite{lin2024awqactivationawareweightquantization} quantized version of these models to save GPU memory and utilize the vLLM~\cite{kwon2023efficient} framework for efficient inference.
Regarding the decoding algorithm, we follow their official guideline\footnote{\url{https://huggingface.co/Qwen/QwQ-32B}} and use greedy decoding to generate the outputs, where temperature is set to 0.7 and top-p is set to 0.95.
In the efficiency and utility experiment, the max output token number of the AIME-2024 dataset is set to 30,000 due to the problem complexity, while we set the max output token number to 20,000 for the remaining datasets. 

\mypara{Baseline}
\begin{itemize}[leftmargin=*,itemsep=0.1em, topsep=0.2em]
\item{\textit{Tokenskip}~\cite{xia2025tokenskip}} first constructs a dataset where less important tokens are pruned and fine-tunes the models on the compressed dataset to enable the model to selectively skip the redundant tokens in inference. 
For GSM-8k and MATH-500, we first fine-tune the model on the training set and evaluate the performance on the test set, where the training dataset has 7,453 and 7,500 problems, respectively.
Since AIME-2024 and Code have no official training set, we transfer the tuned model from MATH, which includes more challenging problems.

\item \textit{CoT-Valve}~\cite{ma2025cot} utilizes interpolation of the LLMs' and LRMs' parameters to collect CoTs of varying lengths, followed by progressively fine-tuning the LRMs to compress the length of CoT.
We fine-tune the model on their officially provided dataset, i.e., MixChain-Z-GSM8K (6,863 samples), and select the best model for comparison.
Specifically, we choose CoT-Valve+P as the fine-tuning pattern.
\end{itemize}
\subsection{Suboptimal Mode}
\mypara{Flawed Thought Identification}
\begin{cotbox}[Flawed Thought]
    You are a highly skilled mathematical reasoning evaluator. Your task is to do the following:

    1. **Analyze a provided solution or reasoning** for the same problem.
    
    2. **Check whether the reasoning is correct**, identifying any logical flaws, computational errors, or misconceptions.
    
    3. Specifically:
    - Verify the mathematical validity of each step.
    
    - Determine if the final answer matches the correct one.
    
    - Explain clearly whether the idea or solution is correct or not, and why.

    Be thorough, precise, and educational in your response

    Problem: {problem}
    Reasoning: {reasoning}
    
    Your task:
    1. Evaluate the reasoning and answer above. Is the logic correct? If not, where and why does it fail? 
    
    2. You don't need to provide a correct answer or provide a detailed analysis.
    
    3. If any incorrect part exists, output True enclosed by boxed, otherwise False.
\end{cotbox}
\begin{cotbox}[Sample Output of Flawed Thought]
\textbf{\color{highlight}[Input Question]}
In triangle $ABC$, $AB = 17$, $AC = 8$, and $BC = 15$.  Let $D$ be the foot of the altitude from $C$ to $AB$.  Find the area of triangle $ACD$.

    \textbf{\color{highlight}[Input Resoning]}
To find the area of triangle $ACD$, we can use the following key points:

1. Recognize that $D$ is the foot of the altitude from $C$ to $AB$, which means $CD$ is perpendicular to $AB$.

2. Use Heron's formula or another method to calculate the area of the whole triangle $ABC$, as it will be useful for later steps.

3. The area of triangle $ABC$ can also be expressed as $\frac{1}{2} \times \text{base} \times \text{height}$, where the base is $AB$ and the height is $CD$.

4. Since we have the length of $AB$ and can find the area of triangle $ABC$, we can solve for the length of $CD$.

5. Knowing $CD$ (the height) and $AC$ (one of the bases), we can then find the area of triangle $ACD$ using the formula $\frac{1}{2} \times \text{base} \times \text{height}$.

6. It's important to note that since $D$ is on $AB$, $AD + DB = AB$, but we don't need this information directly to find the area of $ACD$.
    \tcblower
    \textbf{\color{highlight}[Error Analysis]}
We are given a problem involving triangle $ABC$ with side lengths:

- $AB = 17$
- $AC = 8$
- $BC = 15$

Point $D$ is the foot of the altitude from $C$ to $AB$, and we are to **find the area of triangle $ACD$**.

---

Step 1: Analyze the provided reasoning

The reasoning outlines the following steps:

1. $D$ is the foot of the altitude from $C$ to $AB$, so $CD \perp AB$.  (Correct)

2. Use Heron’s formula or another method to compute the area of triangle $ABC$.  (Valid idea)

3. The area of triangle $ABC$ can also be written as $\frac{1}{2} \cdot AB \cdot CD$, since $CD$ is the height from $C$ to base $AB$.  (Correct)

4. Since $AB = 17$ and area of $ABC$ can be computed, solve for $CD$.  (Valid)

5. Then, **use $AC$ as a base and $CD$ as the height** to compute the area of triangle $ACD$ using $\frac{1}{2} \cdot AC \cdot CD$.  (**This is incorrect**)

6. Notes that $AD + DB = AB$, but claims it's not needed. (Partially true, but missing insight)

---

Step 2: Identify the flaw

The **critical error** lies in **Step 5**:

> "Knowing CD (the height) and AC (one of the bases), we can then find the area of triangle ACD using the formula 1/2 * base * height."

This assumes that $CD$ is the height **relative to base $AC$** in triangle $ACD$. But **this is not true**.

- In triangle $ACD$, if we take $AC$ as the base, then the height must be the perpendicular distance from $D$ to line $AC$.

- However, $CD$ is perpendicular to $AB$, **not** to $AC$.

- Therefore, $CD$ is **not** the height corresponding to base $AC$ in triangle $ACD$.

Thus, computing $\frac{1}{2} \cdot AC \cdot CD$ does **not** give the area of triangle $ACD$. That formula would only be valid if $CD$ were perpendicular to $AC$, which it is not.

Even though $CD$ is an altitude to $AB$, it is **not** an altitude to $AC$, so using it as the height with base $AC$ is geometrically invalid.

---

Step 3: Conclusion

The logic fails in Step 5 due to a **misapplication of the area formula**. The height $CD$ cannot be used with base $AC$ to compute the area of triangle $ACD$, because the height must be measured perpendicular to the chosen base.

While earlier steps (computing area of $ABC$, solving for $CD$) are valid and useful, the final step incorrectly assumes $CD$ serves as a height relative to $AC$, which it does not.

To correctly find the area of triangle $ACD$, one would need to find either:

- The length $AD$, then use $\frac{1}{2} \cdot AD \cdot CD$ (since $CD \perp AB$, and $AD$ lies on $AB$), or

- Use coordinates or trigonometry to find the appropriate height.

---

Final Answer:

$$
\boxed{\text{True}}
$$

The reasoning contains an error — specifically, the incorrect assumption that $\frac{1}{2} \cdot AC \cdot CD$ gives the area of triangle $ACD$. Therefore, the answer is **True**, indicating that the reasoning **is flawed**.

\end{cotbox}

\mypara{Thought Deviation}
\begin{cotbox}[Prompt For Identifying Thought Deviation]
    Here are the reasoning steps provided by the model:
    
        {segments}

        And here is the reference CoT:
        
        {reference\_cot}

        Please analyze the following:
        
        1. How many steps does the reference CoT contain?
        
        2. How many steps of the reference CoT are followed by the model’s response?
        
        3. Does the model adopt any new approach or steps that were not mentioned in the reference CoT?

        Output:
        
        - Number of reference CoT steps: [number of steps]
        
        - Percentage of followed steps: [percentage \%]
        
        - Does the model adopt a new way to solve the 
        problem: [Yes/No]
\end{cotbox}
\begin{cotbox}[Sample Output of Deviation - 1]
- Number of reference CoT steps: 6

- Percentage of followed steps: 100\%

- Does the model adopt a new way to solve the problem: Yes

The model's response follows all the steps outlined in the reference CoT, but it also includes additional details and calculations, such as explicitly checking specific values of ( k ) and verifying the calculations for ( m ) and ( n ). 

This indicates that the model adopts a more detailed approach to ensure the solution is correct, which was not explicitly mentioned in the reference CoT."
\end{cotbox}

\begin{cotbox}[Sample Output of Deviation - 2]
- Number of reference CoT steps: 4

- Percentage of followed steps: 0\%

- Does the model adopt a new way to solve the problem: Yes

The model's response takes a different approach by analyzing the parity of each intermediate function in the composition, rather than focusing on the sign changes with each application of f and g as in the reference CoT.
\end{cotbox}
\subsection{Identifying the Skip of Thought}
Our initial experiments on MATH-500 show that a simple heuristic—checking for the absence of phrases like "Alright" or "I think"—can detect skipped reasoning with about 90\% recall, meaning it correctly identifies most cases where the model skips internal thinking. 
However, the precision for detecting rethinking (i.e., when the model does not skip) is only around 40\%, meaning many predicted rethinking cases are false positives. 

This low precision is not a major concern because rethinking is a useful and often necessary behavior. 
When the external CoT is incomplete or wrong, the LRMs needs to generate its own reasoning to arrive at the correct answer. 
While this increases token usage, it ensures accuracy. 
Therefore, high recall in detecting skipping is more important—it lets us reliably identify when the model is saving computation, while occasional misclassification of rethinking does not harm overall performance or efficiency. 
We leave more accurate detection methods for future work.

\subsection{Generalization Ability}
To evaluate the generalization capability of our approach across different task domains, we conduct experiments on the Code dataset, which involves code generation tasks fundamentally distinct from mathematical reasoning. 
We investigate whether task-specific prompting is necessary for effective performance or if a general-purpose CoT generation framework can transfer seamlessly. 

\mypara{Prompt for Code Reasoning}
The prompt for generating code-specific CoT is shown as follows:

\begin{cotbox}[Example of Code-specific CoT]
``
1. Initialize a counter to zero, which will keep track of the number of operations.\newline
2. Iterate through the array `nums` and count how many elements are less than `k`.\newline
3. Each element less than `k` represents one operation needed to remove it.\newline
4. Return the total count as the minimum number of operations required.\newline
5. If all elements are already greater than or equal to `k`, return 0 operations.\newline<STOP>''
\end{cotbox}

\mypara{Difference in Generated CoTs}
We provide two examples to demonstrate the difference between the CoTs generated by general and code-specific prompts.
The code-specific prompts generally provide more details and reminders for the code-writing task.
\begin{cotbox}[Example of General CoT]
``To solve this problem, follow these key points: \ newline
1. Understand that the goal is to remove elements smaller than `k` until all remaining elements are greater than or equal to `k`.\newline
2. Count how many elements in the array are less than `k`.\newline
3. The count of such elements will be the minimum number of operations required.\newline<STOP>''

\end{cotbox}

In the baseline setting, we employ the standard, domain-agnostic prompt described in previous sections.
At the same time, we further test a specialized variant of this prompt—adapted to emphasize code-related reasoning (e.g., algorithmic structure and control flow), while still adhering to the principle of avoiding implementation-level details. 

The results, shown in \Cref{tab:dataspecific}, reveal no significant performance difference between the two settings. 
This finding highlights the one-for-all property of our approach: the general CoT generation framework is robust enough to handle diverse tasks without requiring task-specific adjustments. 
In other words, \Method demonstrates strong adaptability across domains, eliminating the need for choosing different CoT templates for different types of problems. 

\begin{table}[t]
\centering
\caption{\textbf{Generalization Ability}: The results are reported using Pass@1 and the number of generated tokens. Normal represents using original COTs, while Specific represents using the task-specific CoTs.  }
\label{tab:dataspecific}
\resizebox{\linewidth}{!}{
\begin{tabular}{c|ccccc|ccccc}
\toprule
 & \multicolumn{5}{c|}{QwQ-32B} & \multicolumn{5}{c}{DPSK-32B} \\ \cmidrule{2-11} 
 & \multicolumn{2}{c}{Normal} &  & \multicolumn{2}{c|}{Specific} & \multicolumn{2}{c}{Normal} &  & \multicolumn{2}{c}{Specific} \\ \cmidrule{2-3} \cmidrule{5-6} \cmidrule{7-8} \cmidrule{10-11} 
\multirow{-3}{*}{Ex.} & Pass@1 & Tokens &  & Pass@1 & Tokens & Pass@1 & Tokens &  & Pass@1 & Tokens \\ \midrule
 3b & 65.6 & 6842 &  & 64.1 & 7009 & 59.1 & 6557 &  & 55.7 & 6163 \\
 7b & 62.3 & 4409 &  & 61.7 & 4485 & 41.7 & 528 &  & 42.4 & 627 \\
 Plus & 64.1 & 4461 &  & 61.9 & 4408 & 45.8 & 528 &  & 45.3 & 539 \\
Max & 60.9 & 4209 &  & 62.2 & 4181 & 44.5 & 465 &  & 41.1 & 577 \\
\bottomrule
\end{tabular}
}
\end{table}

\subsection{Performance under Different Token Budgets}
\begin{figure}[t!]
    \centering
    \includegraphics[width=1\linewidth]{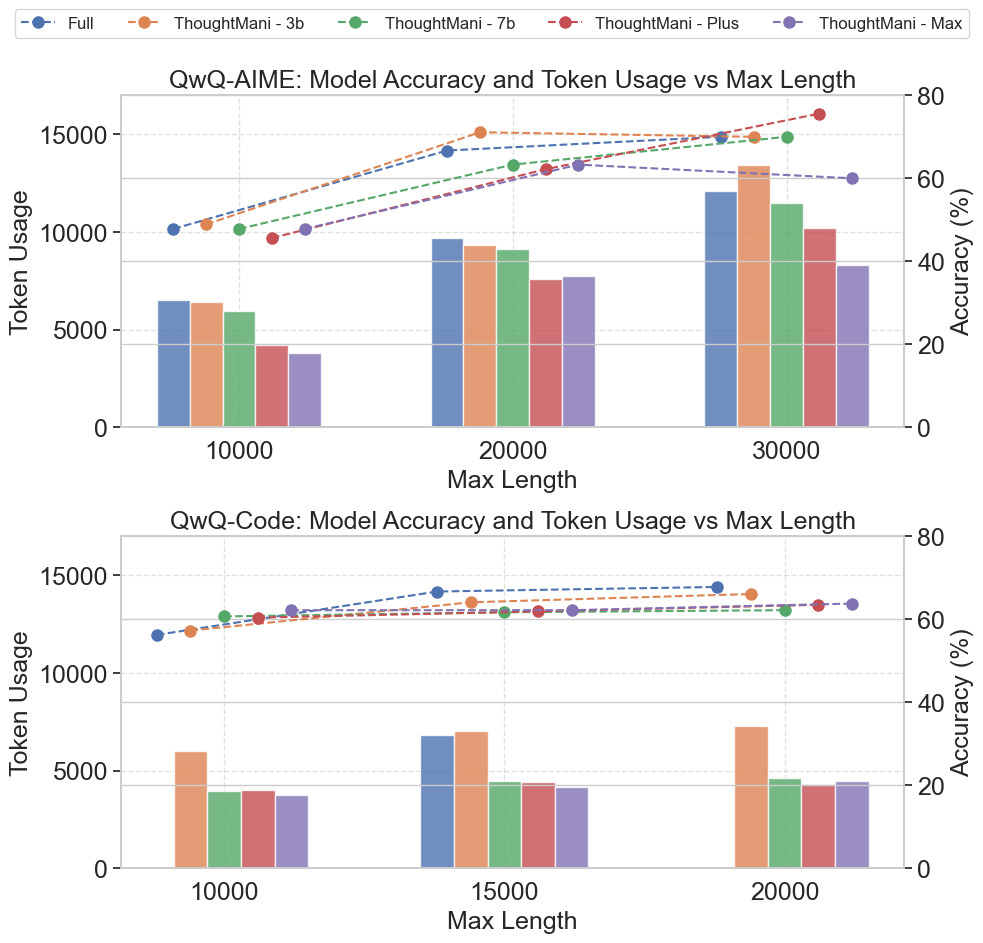}
    \caption{\textbf{Relation between Model Performance and Token Budgets:} Bar plot represents the token consumption and line plot represents the model utility (Accuracy or Pass@1)}
    \label{fig:budget}
\end{figure}

For more challenging datasets, such as AIME and Code, the inference process of RL-based LRMs typically demands a significant number of tokens to achieve high-quality reasoning. 
To systematically study the impact of token budgets on performance, we adjust the max\_model\_length parameter during inference and evaluate the corresponding outcomes. 
Specifically, for the AIME dataset, we set the maximum token length to 10,000, 20,000, and 30,000, while for the Code dataset, we use 10,000, 15,000, and 20,000. 
The results are presented in \Cref{fig:budget}.
Generally, as the max\_model\_length increases, the accuracy improves while maintaining inference efficiency. 

An interesting observation is that the performance gap between the baseline and \Method is more pronounced when the token budget is limited. 
One possible explanation is that the externally generated CoTs, especially for very hard problems, may include hallucinations or insufficient information. 
In such cases, the LRM compensates by generating additional thoughts to supplement the provided CoTs, showcasing one limitation of \Method. 
In contrast, for simpler datasets like GSM-8k, where most questions can be resolved within fewer than 1,000 tokens, this performance gap does not exist. 
These findings underscore the importance of balancing token budgets with problem complexity. 
While \Method demonstrates benefits in reducing computational overhead, it shows effectiveness more evidently in scenarios where the token budget is sufficient.

\end{document}